# MAP Estimation of Semi-Metric MRFs via Hierarchical Graph Cuts

**M. Pawan Kumar**
Computer Science Department
Stanford University
pawan@cs.stanford.edu

**Daphne Koller**
Computer Science Department
Stanford University
koller@cs.stanford.edu

## Abstract

We consider the task of obtaining the *maximum a posteriori* estimate of discrete pairwise random fields with arbitrary unary potentials and semi-metric pairwise potentials. For this problem, we propose an accurate hierarchical move making strategy where each move is computed efficiently by solving an $st$-MINCUT problem. Unlike previous move making approaches, e.g. the widely used $\alpha$-expansion algorithm, our method obtains the guarantees of the standard linear programming (LP) relaxation for the important special case of metric labeling. Unlike the existing LP relaxation solvers, e.g. interior-point algorithms or tree-reweighted message passing, our method is significantly faster as it uses only the efficient $st$-MINCUT algorithm in its design. Using both synthetic and real data experiments, we show that our technique outperforms several commonly used algorithms.

## 1 Introduction

Markov random fields (MRFs) offer an expressive and intuitive framework for several important problems in artificial intelligence and machine learning. Given a set of random variables along with a neighborhood relationship defined over them, an MRF offers a concise representation of the probability of each labeling (i.e. a particular assignment of labels to the variables) in terms of potentials defined over the cliques of random variables. Due to the central role of MRFs in various applications, algorithms that perform efficient and accurate inference on them are highly desirable. One important and well-studied class of inference, called *maximum a posteriori* (MAP) estimation, seeks the labeling with the maximum probability.

We consider a special case of MAP estimation, known as semi-metric labeling [4], where (i) the size of the maximal clique is 2 (a pairwise MRF); and (ii) the pairwise potentials are defined by a semi-metric distance function over labels. Although these may seem like very restrictive assumptions, several problems in computer vision and related areas can be expressed using semi-metric labeling, from low level tasks like image denoising and stereo reconstruction [30] to high level tasks like pose estimation [9] and scene segmentation [27]. Hence, the semi-metric labeling problem merits special attention.

We describe a novel algorithm for semi-metric labeling which approximates a given semi-metric distance function using a mixture of *r-hierarchically well-separated tree* ($r$-HST) metrics [1]. The $r$-HST metrics form an amenable class of distance functions which admit elegant divide-and-conquer approaches for several problems [1, 8]. In our case, they not only result in easier-to-solve instances of MAP estimation, they also provide an accurate approximation of the original problem. Given a mixture of $r$-HSTs, we reformulate semi-metric labeling using a set of $r$-HST *metric labeling problems* (i.e. MAP estimation for $r$-HST metric pairwise potentials), where each problem is specified by one component of the mixture. We show how each resulting $r$-HST metric labeling problem can be solved accurately using an iterative procedure that only employs the efficient $st$-MINCUT algorithm [3] in its design. Unlike previous $st$-MINCUT based approaches, our method provides the best known approximation bound for the important special case of metric labeling (i.e. when the pairwise potentials are defined by a metric distance function). In practice, our technique outperforms several state of the art algorithms on both synthetic and real data experiments.

## 2 Related Work

The most commonly used algorithms for semi-metric labeling can be broadly divided into two categories: message passing and move making. Message passing algorithms attempt to minimize approximations of the free energy associated with the MRF [10, 14, 15, 29, 34, 36]. Amongst them, the algorithms of [14, 15, 34] are closely related to the linear programming (LP) relaxation of semi-metric labeling [7, 17, 25, 33]. Although message passing algorithms provide accurate MAP estimates, they can be computationally expensive in certain cases [30].

Move making approaches refer to a large class of iterative algorithms which *move* from one labeling to the other while ensuring that the probability of the labeling never decreases. The move space, i.e. the search space for the new labeling, is restricted to a subspace of the original search space that can be explored efficiently. Typically, the move space is explored using the $st$-MINCUT algorithm, e.g. in $\alpha\beta$-swap [4] and $\alpha$-expansion [5]. However, recently researchers have also used more sophisticated algorithms such as quadratic pseudo-boolean optimization [2] (e.g. see [13, 21, 35]). Move making algorithms are generally preferred in applications which involve a large number of random variables, e.g. on image-sized MRFs in computer vision, due to their efficiency.

## 3 Preliminaries

Consider an MRF defined over a set of random variables $\mathcal{V} = \{v_1, v_2, \cdots, v_N\}$, each of which can take a value from a discrete label set $\mathcal{L} = \{l_1, l_2, \cdots, l_H\}$. Furthermore, let $\mathcal{E}$ define the neighborhood such that $v_a$ and $v_b$ are neighbors if and only if $(v_a, v_b) \in \mathcal{E}$. A labeling of the MRF is a function $f : \{1, 2, \cdots, N\} \rightarrow \{1, 2, \cdots, H\}$ such that variable $v_a$ takes label $l_{f(a)}$. Associated with each labeling is its probability $\Pr(f|\boldsymbol{\theta}) = \exp(-Q(f; \boldsymbol{\theta}))/Z$, where $Z$ is the partition function and $Q(\cdot; \boldsymbol{\theta})$ is the Gibbs energy:

$$Q(f; \boldsymbol{\theta}) = \sum_{v_a \in \mathcal{V}} \theta_a(f(a)) + \sum_{(v_a, v_b) \in \mathcal{E}} \theta_{ab}(f(a), f(b)). \quad (1)$$

Here $\theta_a(f(a))$ and $\theta_{ab}(f(a), f(b))$ denote unary and pairwise potentials respectively. For semi-metric labeling, the pairwise potentials are of the form $\theta_{ab}(f(a), f(b)) = w_{ab} d(f(a), f(b))$, where $w_{ab} \geq 0$ and $d(\cdot, \cdot)$ is a semi-metric distance function. Recall that $d(\cdot, \cdot)$ is semi-metric if and only if: (i) $d(i, i) = 0, \forall i$; and (ii) $d(i, j) = d(j, i) > 0, \forall i \neq j$. Examples of commonly used semi-metric distance measures include the truncated linear function, $d(i, j) = \min\{|i - j|, M\}$ where the truncation factor $M \geq 0$, the truncated quadratic function, $d(i, j) = \min\{(i - j)^2, M\}$, and the uniform metric (a special case of truncated linear/quadratic function with $M = 1$). Within this setting, the problem of MAP estimation is formally specified as: $f^* = \arg\min_f Q(f; \boldsymbol{\theta})$.

## 4 The $r$-HST Metric Labeling Problem

As mentioned earlier, there are two key ingredients to our MAP estimation algorithm: (i) approximating a given semi-metric by a mixture of $r$-HST metrics; and (ii) solving each resulting $r$-HST metric labeling problem. We begin by defining $r$-HST metrics and designing an efficient move making algorithm for the corresponding labeling problem. The next section describes a simple yet accurate procedure for approximating semi-metrics.

### 4.1 The $r$-HST Metric

An $r$-HST metric [1] $d^t(\cdot, \cdot)$ is specified by a rooted tree whose edge lengths are non-negative and satisfy the following properties: (i) the edge lengths from any node to all of its children are the same; and (ii) the edge lengths along any path from the root to a leaf decrease by a factor of at least $r > 1$. Given such a tree, known as $r$-HST, the distance $d^t(i, j)$ is the sum of the edge lengths on the unique path between them. Note that, as the name suggests, an $r$-HST specifies a metric distance. In other words, it is a semi-metric distance function that satisfies the triangular inequality: $d(i, j) - d(j, k) \leq d(i, k), \forall i, j, k$. In this paper, we consider only those $r$-HSTs where all the labels in the set $\mathcal{L}$ are at the leaves of the $r$-HST. As observed in several earlier works [1, 6, 8], $r$-HSTs satisfying this assumption are sufficient to provide an accurate approximation of a given semi-metric distance function. Fig. 1 shows an example $r$-HST over $H = 6$ labels with $r = 2$.

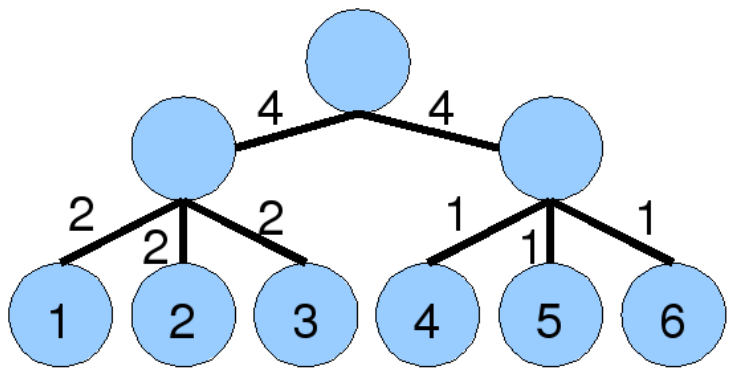

Figure 1: *An example $r$-HST metric. The distances between nodes are specified by path lengths, e.g. $d^t(1, 2) = 4$, $d^t(4, 5) = 2$ and $d^t(1, 4) = 11$.*

### 4.2 The Move Making Algorithm

For a given $r$-HST metric $d^t(\cdot, \cdot)$, we define an MRF parameterized by $\boldsymbol{\theta}^t$ with arbitrary unary potentials $\theta^t_a(i)$ and pairwise potentials of the form $\theta^t_{ab}(i, j) = w_{ab} d^t(i, j)$. We show how to obtain an accurate MAP estimate for the parameter $\boldsymbol{\theta}^t$, known as the $r$-HST metric labeling problem, using a novel approach based on the $st$-MINCUT algorithm. Our approach is a divide-and-conquer method consisting of two steps: (i) replace the original problem by a series of subproblems that are easier to solve; and (ii) combine the solutions of the subproblems to obtain an accurate solution of the original problem. Each of the subproblems is specified by a node in the given $r$-HST, and their solutions are combined using the standard $\alpha$-expansion algorithm [5].

In more detail, consider a node $p$ of the given $r$-HST. We say that a label $l_i$ *belongs to* the node $p$ (denoted by $i \in p$) if and only if it is a leaf node in the subtree rooted at $p$. Let $\mathcal{L}^p$ denote the set of labels that belong to $p$, i.e. $\mathcal{L}^p = \{i | i \in p\}$. The subproblem defined at node $p$ is to find the labeling $f^p$ of the random variables $\mathcal{V}$ that minimizes the energy under the constraint that each variable $v_a \in \mathcal{V}$ takes a label from the set $\mathcal{L}^p$. Note that if $p$ is the root node of the given $r$-HST, then the subproblem is the same as the original problem. On the other hand, if $p$ is the leaf node, then the solution to the subproblem is trivial, i.e. $f^p(a) = p$ for all $v_a \in \mathcal{V}$. In fact, as one moves from the root towards the leaves of the $r$-HST, the label set of the subproblem keeps reducing in size thereby making the subproblems easier to solve. This observation suggests the following hierarchical approach: *solve the easier subproblems at level $m + 1$ of the $r$-HST and use their labelings to solve the subproblem defined by their parent nodes at level $m$.*

It remains to be seen how exactly a subproblem at node $p$ can benefit from the labelings of its child nodes. To answer this we consider the stage of the above hierarchical approach where we have to solve the subproblem defined at node $p$, having already obtained the labelings of the subproblems associated with the children of $p$. We denote these labelings by $f_1, \cdots, f_C$ where $C$ is the number of child nodes of $p$. To find the labeling $f^p$ for the subproblem at $p$ efficiently, we restrict the label of each variable $v_a$ to be one of the $C$ labels specified by the child nodes. In other words, $f^p(a) = f_i(a)$ where $i \in \{1, \cdots, C\}$ is the index of the child node from which $v_a$ takes its label. Note that different variables can take labels from different child nodes. In order to find the indices of the child nodes for all variables, we define a parameter $\boldsymbol{\theta}^p$ such that the corresponding unary and pairwise potentials are given by

$$\theta^p_a(i) = \theta^t_a(f_i(a)), \quad \theta^p_{ab}(i,j) = w_{ab} d^t(f_i(a), f_j(b)),$$
$$\forall (v_a, v_b) \in \mathcal{E}, i, j \in \{1, \cdots, C\}. \qquad (2)$$

We obtain an approximate MAP estimate $f'$ for the parameter $\boldsymbol{\theta}^p$ using $\alpha$-expansion (see [18] for details). Using the properties of $r$-HST metrics, it follows that each move of $\alpha$-expansion results in a submodular problem that can be solved exactly. The labeling $f'$ provides the required indices of the child nodes to obtain the labeling $f^p$ as

$$f^p(a) = f_i(a) \text{ where } i = f'(a), \forall v_a \in \mathcal{V}. \qquad (3)$$

The hierarchical approach for solving the $r$-HST metric labeling problem terminates when the subproblem corresponding to the root node is solved. Our method is not only easy to implement and effective in practice, it also provides the approximation bounds of the LP relaxation [7, 11]. Specifically, the following property holds true:

**Theorem 4.1 [18]:** For $r$-HST metric labeling we obtain an approximation bound of $O(1)$.

## 5 The Semi-Metric Labeling Problem

Similar to MAP estimation, several problems specified on $r$-HST metrics are well-known to be amenable to efficient divide-and-conquer approaches [1, 8]. However, their use in the AI community has been very limited thus far. The main reason for this would appear to be their restrictive form which may not offer an accurate model for real-world applications. A natural way to address this deficiency is to use a mixture of $r$-HST metrics instead of a single $r$-HST.

### 5.1 Learning a Mixture of $r$-HSTs.

Given a distance function $d(\cdot, \cdot)$, we would like to learn a set of $r$-HST metrics $\mathcal{D} = \{d^t(\cdot, \cdot), t = 1, \cdots, T\}$ along with a probability distribution $\boldsymbol{\rho}$ on them such that the *distortion* is minimized, that is

$$(\mathcal{D}^*, \boldsymbol{\rho}^*) = \arg\min_{\mathbf{D}, \boldsymbol{\rho}} \left( \max_{i \neq j} \frac{\sum_t \rho^t d^t(i,j)}{d(i,j)} \right). \qquad (4)$$

When the distance function is a metric, Fakcharoenphol *et al.* [8] provide a simple yet accurate randomized algorithm for sampling $r$-HST metrics. Below, we describe their method for $r = 2$ while noting that it can be easily extended for any value of $r$.

It is helpful to think of each level of an $r$-HST as a clustering of labels such that a node $p$ defines a cluster of labels $\mathcal{L}^p = \{i | i \in p\}$ (i.e. $i$ is a leaf node of the subtree rooted at $p$). In other words, an $r$-HST defines a hierarchical clustering of labels. Let the clustering at level $m$ be denoted by $\mathcal{C}^m$. The root node denotes the trivial clustering which consists of all the labels $\mathcal{C}^1 = \{1, \cdots, H\}$. Given the clusters $\mathcal{C}^{m-1}$, $\mathcal{C}^m$ is obtained by further clustering the labels $\{j | j \in p\}$ for each node $p \in \mathcal{C}^{m-1}$.

Without loss of generality, we assume that the diameter $\Delta = \max_{i \neq j} d(i,j) = 2^\delta$ for an integer $\delta \geq 1$ and $\min_{i \neq j} d(i,j) > 1$. Due to the above assumptions, the $r$-HST would consist of at most $\delta$ levels. The algorithm is initialized by: (i) picking a random permutation $\pi$ of the label indices $\{1, 2, \cdots, H\}$, which defines a priority ordering on the cluster centers; and (ii) choosing a value of $\beta \in [1, 2]$ from the distribution $\Pr(x) = 1/(x \log 2)$. Note that both the permutation $\pi$ and value of $\beta$ are fixed throughout the process, i.e. they are selected once before running the clustering algorithm for all levels. Given $\mathcal{C}^{m-1}$, the clustering at level $m$ is obtained as follows:

- Consider a node $p \in \mathcal{C}^{m-1}$. Define $\beta_m = 2^{\delta - m}\beta$.
- For a label $i \in p$, find the first label $j$ according to the permutation $\pi$ such that $d(i,j) \leq \beta_m$.
- Assign the label $i$ to the cluster centered at $j$.
- Repeat for all labels $i \in p$ and nodes $p \in \mathcal{C}^{m-1}$.

The edge length $e^p$ from a node $p$ to each of its children is given by $\Delta^p/2$ where $\Delta^p$ denotes the diameter of the cluster of labels $\mathcal{L}^p$ specified by $p$. Fakcharoenphol *et al.* [8] showed that $\Delta^p$ reduces by at least a factor of $r = 2$ for metric distances, thereby providing us with an $r$-HST.

Importantly, the method of [8] can also be applied for approximating semi-metric distance functions. However, as the triangular inequality is not satisfied, the resulting tree will not be an $r$-HST. Nonetheless, the tree obtained using this method would provide a metric distance function which can then be approximated to $r$-HST metrics by applying the above procedure again. The only question that remains is the number of $r$-HSTs $T$ to be employed. In order to answer this question, we note that [8] also provided a deterministic version of the above algorithm for solving a related problem that we call the dual procedure (DP):

$$\text{DP}(\mathbf{y}): \min_{d^t(\cdot,\cdot)} \sum_{i,j} y_{ij} d^t(i,j), \text{ s.t. } d^t(i,j) \geq d(i,j), \qquad (5)$$

for some values of $y_{ij} \geq 0$. In other words, DP provides one $r$-HST that minimizes the non-negatively weighted sum of distances that *dominate* the original distance $d(\cdot, \cdot)$ (i.e. $d^t(i,j) \geq d(i,j)$, for all $i, j$). Note that each $r$-HST sampled from the randomized procedure described above dominates $d(\cdot, \cdot)$. Briefly, DP works by derandomizing the above procedure using conditional expectation. In other words, the elements of the permutation $\pi$ are obtained sequentially by computing the expectation of equation (5) given the previously selected elements of the permutation. Using DP, Charikar *et al.* [6] provided an iterative algorithm to obtain a small mixture of $r$-HST metrics (with $O(H \log H)$ $r$-HSTs). The algorithm initializes the mixture to one $r$-HST. In our implementation, we found the $r$-HST obtained by solving the DP (5) for the values $y_{ij} = 1$ for all $i, j$ to be a good initialization. Let $(\mathcal{D}, \boldsymbol{\rho})$ denote the mixture after $n$ iterations which consists of $n$ $r$-HSTs. The $(n+1)^{st}$ $r$-HST is obtained by defining

$$y_{ij} = \begin{cases} \frac{1}{d(i,j)} \exp\left(\frac{\sum_t \rho^t d^t(i,j)}{\lambda d(i,j)}\right) & \text{if} \quad i \neq j, \\ 0 & \text{otherwise}, \end{cases} \tag{6}$$

and solving DP(**y**). Here $\lambda > 0$ is the user defined learning rate. Note that in the above values of $y_{ij}$, the pairs of labels $l_i$ and $l_j$ which result in a bigger distortion are given more weight while solving the DP. The probability distribution over the $n+1$ $r$-HSTs is updated to $((1-\lambda)\boldsymbol{\rho}; \lambda)$ where $(;)$ denotes vector concatenation. Although more sophisticated clustering algorithms may be used, the above method is appealing due to its ease of implementation. Furthermore, it also provides an accurate approximation as evidenced by the following result from [8] and its simple extension.

**Theorem 5.1 [8]:** When $d(\cdot,\cdot)$ is a metric distance function, the above approach provides a mixture of $r$-HST metrics with distortion of $O(\log H)$.

**Theorem 5.2:** Let $d(\cdot,\cdot)$ be a semi-metric which satisfies the following relaxed version of triangular inequality:

$$d(i,j) - d(j,k) \leq \gamma d(i,k), \forall i,j,k, \tag{7}$$

for some value of $\gamma \geq 1$. The above approach provides a mixture of $r$-HST metrics with a distortion of $O((\gamma \log H)^2)$ with respect to $d(\cdot,\cdot)$. Note that any distance function defined over a finite number of labels will admit a finite $\gamma$.

### 5.2 Approximating Semi-Metric Labeling

Once the mixture of $r$-HSTs is learnt, the original semi-metric labeling problem parameterized by $\boldsymbol{\theta}$ can be approximated by a set of $r$-HST metric labeling problems specified by parameters $\boldsymbol{\theta}^t, t = 1, \cdots, T$, where

$$\theta^t_a(i) = \theta_a(i), \quad \theta^t_{ab}(i,j) = w_{ab} d^t(i,j). \tag{8}$$

As shown in the previous section, $r$-HST metric labeling can be solved efficiently and accurately using an $st$-MINCUT based approach. Hence, in order to solve semi-metric labeling, we solve the set of $r$-HST metric labeling problems to obtain the labelings $f^t$. We then combine these labelings to obtain the final labeling $f$, using the same approach as the one used to combine the labelings of the children of node $p$ of the $r$-HST (see § 4.2). Note that, unlike the problem of combining labelings of child nodes, in this case the moves of $\alpha$-expansion are no longer necessarily submodular. In other words, we are not guaranteed to obtain the optimal move at each iteration. In order to overcome this problem, we solve the $\alpha$-expansion procedure by using the primal dual scheme of [16]. This has two advantages: (i) it reduces the run-time of $\alpha$-expansion; and (ii) it handles non-submodular moves by truncating the edges with negative capacities in the $st$-MINCUT graph to 0. At each iteration of $\alpha$-expansion, we move to a new labeling only if it decreases the energy. Otherwise we retain the old labeling and repeat the procedure until we can no longer decrease the energy for any iteration of $\alpha$-expansion. We initialize the labeling by the lowest energy labeling amongst the set $\{f^t, t = 1, \cdots, T\}$. The $\alpha$-expansion procedure described above guarantees that the energy is not increased at any iteration. In other words, the energy of the labeling obtained by our approach is bounded from above by the energy of the best labeling provided by solving the set of $r$-HST metric labeling problems. Using this observation along with Theorems 5.1 and 5.2 allows us to prove the following approximation bounds for our overall approach.

**Theorem 5.3 [18]:** For the metric and semi-metric labeling problems, we obtain an approximation bound of $O(\log H)$ and $O((\gamma \log H)^2)$ respectively.

In practice, when solving a subproblem at node $p$ of an $r$-HST, we use the given distance function $d(\cdot,\cdot)$ to specify the pairwise potentials of $\boldsymbol{\theta}^p$ instead of $r$-HST metric $d^t(\cdot,\cdot)$. This tends to improve the quality of the labelings whilst retaining the approximation bound. Specifically,

**Observation 5.4 [18]:** Theorem 5.3 also holds true if the $r$-HST metric $d^t(\cdot,\cdot)$ is replaced by the given distance $d(\cdot,\cdot)$ in equation (2) for all subproblems defined by the $r$-HSTs.

Note that our algorithm provides the guarantees of the LP relaxation for the metric labeling problem. Together with the results for truncated convex models [5, 20], this implies that there exist moving making algorithms which match all known LP relaxation guarantees when the number of labels is smaller than the number of variables (i.e. $H < N$). Although the above theorem shows that our approach provides a tight approximation, we can further improve its accuracy by using a hard EM strategy described below.

### 5.3 Refining the Labeling

For a given semi-metric labeling problem, consider the labeling $f$ obtained using the method described above. The energy defined by $f$ is given by

$$Q(f;\boldsymbol{\theta}) = \sum_{v_a \in \mathcal{V}} \theta_a(f(a)) + \sum_{(v_a,v_b) \in \mathcal{E}} w_{ab} d(f(a), f(b)) \tag{9}$$

We define a set of non-negative weights **y** as

$$y_{ij} = \sum_{(v_a,v_b) \in \mathcal{E}, f(a)=i, f(b)=j} w_{ab}, \tag{10}$$

| | (i) | (ii) | (iii) | (iv) | (v) |
|---|---|---|---|---|---|
| $\alpha$-exp | 48645 | 52094 | 50221 | 48112 | 47613 |
| $\alpha\beta$-swap | 48721 | 51938 | 51055 | 48487 | 47579 |
| TRW-S | **47506** | **51318** | **48132** | **47355** | **46612** |
| BP-S | 50942 | 60269 | 52841 | 48136 | 47402 |
| R-swap | 48045 | 51842 | - | - | - |
| R-exp | 47998 | 51641 | - | - | - |
| Our | **47850** | **51587** | **48146** | **47538** | **46651** |
| Our+EM | **47823** | **51413** | **48146** | **47382** | **46638** |

(a) Energy

| | (i) | (ii) | (iii) | (iv) | (v) |
|---|---|---|---|---|---|
| $\alpha$-exp | **0.44** | **0.36** | **0.29** | **0.30** | **0.36** |
| $\alpha\beta$-swap | **0.65** | **0.86** | **0.52** | **0.51** | **0.47** |
| TRW-S | 104.29 | 178.97 | 713.70 | 703.82 | 709.36 |
| BP-S | 15.78 | 45.63 | 150.36 | 129.68 | 141.79 |
| R-swap | **1.97** | **10.73** | - | - | - |
| R-exp | 5.78 | 30.73 | - | - | - |
| Our | 10.22 | 12.84 | **1.86** | **10.58** | **12.25** |
| Our+EM | 25.66 | 64.08 | 5.02 | 32.75 | 57.50 |

(b) Time

Table 1: *Average energy and time (in seconds) of* MAP *estimation algorithms computed using* $100$ *randomly generated* MRFs*. The columns denote the five cases considered (see text). The three smallest average energy values and average timings are shown in bold. Note that range swap and range expansion are only applicable to truncated convex models. Hence, their timing and energy is not reported for other cases. As the results indicate, our approach provides an accurate* MAP *estimate efficiently.*

i.e. $y_{ij}$ is the contribution of labels $l_i$ and $l_j$ to the energy (9). Specifically, using **y**, the energy of the labeling $f$ can be rewritten as

$$Q(f;\boldsymbol{\theta}) = \sum_{v_a \in \mathcal{V}} \theta_a(f(a)) + \sum_{i,j} y_{ij} d(i,j). \tag{11}$$

We obtain an $r$-HST metric $d^t(\cdot,\cdot)$ by solving DP (5) for the values of **y** defined above. The metric $d^t(\cdot,\cdot)$ provides an MRF parameterized by $\boldsymbol{\theta}^t$ as defined in equation (8). Since the metric $d^t(\cdot,\cdot)$ dominates the given distance $d(\cdot,\cdot)$ it follows that

$$\sum_{i,j} y_{ij} d(i,j) \leq \sum_{i,j} y_{ij} d^t(i,j) \Rightarrow Q(f;\boldsymbol{\theta}) \leq Q(f;\boldsymbol{\theta}^t). \tag{12}$$

Now consider the case when the above inequality holds with an equality. In other words, DP provides an $r$-HST metric which exactly models the weighted sum of distances where the weights are specified by **y**. We can now solve the $r$-HST metric labeling problem corresponding to $\boldsymbol{\theta}^t$ in order to obtain a new labeling $f'$. If the labeling $f'$ is such that $Q(f';\boldsymbol{\theta}^t) \leq Q(f;\boldsymbol{\theta}^t)$ then we are guaranteed not to increase the energy of the solution by moving from labeling $f$ to labeling $f'$ since

$$Q(f';\boldsymbol{\theta}^t) \leq Q(f;\boldsymbol{\theta}^t) = Q(f;\boldsymbol{\theta}) \Rightarrow Q(f';\boldsymbol{\theta}) \leq Q(f;\boldsymbol{\theta}). \tag{13}$$

The process of obtaining a new $r$-HST metric followed by a new labeling $f'$ can be repeated till we reach a local minima. Note that the above inequality is obtained by assuming that the DP can be solved exactly. However, this cannot be guaranteed for general semi-metric distance functions. Nonetheless, in practice we use the above procedure to refine the labeling obtained by solving each $r$-HST metric labeling problem. As the results in the next section show, it helps further decrease the energy of the labelings obtained by our method at the cost of more computation time.

## 6 Experiments

We compare our approach to several state of the art MAP estimation algorithms using both synthetic and real data experiments. In all our experiments we set $r = 2$. Empirically, we found that the accuracy of our approach saturates after using $T = 50$ $r$-HSTs to define the mixture.

**Synthetic Data.** We consider the following cases of the MAP estimation problem: (i) truncated linear metrics; (ii) truncated quadratic semi-metrics; (iii) $r$-HST metrics; (iv) general metrics; and (v) general semi-metrics. Note that for uniform metric labeling, our approach reduces to $\alpha$-expansion and hence, we do not consider such problems in our evaluation. In each of the five cases above, we generated $100$ random 4-connected grid structured MRFs of size $100 \times 100$ with $H = 20$. The unary potentials were randomly sampled from the uniform distribution defined over the interval $[0, 10]$ (denoted by $u(0, 10)$). The pairwise potentials for the five cases were generated as follows. For the truncated convex models (cases (i) and (ii)) the truncation factor was sampled from $u(0, 10)$. For $r$-HST metrics we defined a random hierarchical clustering of labels with the edge lengths at the root sampled from $u(0, 10)$. The edge lengths at other levels were sampled while ensuring that the properties of the $r$-HST metric hold true. In order to generate a general metric distance function, we defined a complete graph over the labels with random edge lengths from $u(0, 10)$. The distance function $d(i, j)$ between labels $i$ and $j$ is given by the shortest path from $i$ to $j$. A general semi-metric distance was defined by randomly sampling the values of $d(i, j)$ where $i \neq j$ from $u(0, 10)$ and setting $d(i, i) = 0$ for all $i$.

The MRFs were used to test several state of the art MAP estimation algorithms: $\alpha$-expansion [5], $\alpha\beta$-swap [4], sequential tree-reweighted message passing (TRW-S) [14], sequential belief propagation (BP-S) [24], range swap [32], and range expansion [20]. We used publicly available code for these approaches to compare them with the two variants of our method: with and without using the hard EM strategy described in § 5.3.

The $\alpha$-expansion algorithm was solved using the primal-dual scheme of [16] (for both the original problem as well as the various subproblems used in our approach). Recall that [16] also handles non-submodular moves and hence, is capable of solving semi-metric labeling problems like cases (ii) and (v). All the move making algorithms were initialized to the constant labeling $f(a) = 1$ for all $v_a$. For the truncated convex models (cases (i) and (ii)) the messages of TRW-S and BP-S were computed efficiently using the dis-tance transform technique [9]. We report the results of the methods described in [20, 32] (denoted by R-exp and R-swap respectively) only for truncated convex models since these approaches are not applicable to the other cases.

Table 1 lists the average time required and the average value of the energy obtained for various methods. Our approach is slower than previous move making algorithms ($\alpha$-expansion and $\alpha\beta$-swap) as it solves a set of $r$-HST metric labeling problems. However, in terms of the energy values, it significantly outperforms them in all cases. It even provides similar results to the methods of [20, 32] which were specifically designed for the truncated convex models. The energy values obtained by our approach also compare favorably with TRW-S. In terms of speed, our method is significantly faster than TRW-S, especially in cases where the distance transform trick cannot be employed. As mentioned earlier, the computational efficiency of our method is due to the fact that it only uses the efficient $st$-MINCUT algorithm in its design. Finally, we also note that the hard EM strategy decreases the energy of the labeling. However, it is slower since it has to solve at least one instance of the DP (5) for each $r$-HST in the mixture.

**Scene Registration.** Given two images of different scenes with some common elements (e.g. both scenes contain buildings, see Fig. 2), scene registration requires us to find a point to point correspondence from one image to the other. In this work, we follow the framework of [22] and define an MRF whose variables correspond to the pixels of the first image. The labels of the variables denote the displacement that the pixel undergoes from the first image to its corresponding pixel in the second image. The neighborhood is defined such that the MRF forms a 4-connected grid graph. The unary potentials are given by the $\ell_1$ difference between the SIFT features [23] of corresponding points. The pairwise potentials, which enforce smoothness of the displacement map, are defined as

$$\theta_{ab}(i,j) = \kappa \left( \min\{|u(i) - u(j)|, M\} + \min\{|v(i) - v(j)|, M\} \right), \tag{14}$$

where $(u(i), v(i))$ and $(u(j), v(j))$ are the horizontal and vertical displacements specified by labels $l_i$ and $l_j$ respectively, $M$ is the truncation factor and $\kappa$ is the scaling factor. Since the above pairwise potential forms a metric distance, our approach can be applied to obtain the solution.

In our experiments, we use the values of $u(i) \in [-5, 5]$ and $v(i) \in [-5, 5]$, i.e. the total number of labels for each random variable is $H = 121$. The truncation factor $M$ was set to $5$ and the scaling factor $\kappa = 1$. Fig. 2 shows the results obtained for three pairs of images using six different MAP estimation algorithms along with the corresponding energy values and timings. Similar to the synthetic data experiments, our approach outperforms other move making approaches in terms of accuracy, and it outperforms TRW-S in terms of speed. In fact, the accuracy of our method is very similar to TRW-S. Note that TRW-S and BP-S can be speeded up by using the decomposable model [26]. However, this makes the approximations to the free energy weaker thereby providing less accurate results.

A related problem to scene registration, known as stereo reconstruction, is concerned with obtaining correspondences between two images of the same scene. The image pairs are epipolar rectified, i.e. the vertical displacement of each pixel is known to be $0$. The unary potentials are computed using the difference in the RGB values of the corresponding pixels (instead of the SIFT feature), and the pairwise potentials are given by equation (14) with $M = 5$ and $\kappa = 20$. We compared our approach with other algorithms on two standard stereo pairs used in computer vision, namely 'teddy' and 'tsukuba'. Our method provides a labeling with lower energy than $\alpha$-expansion and $\alpha\beta$-swap using $H = 40$ labels, as shown in Fig. 3.

**Image Denoising** Image denoising is a classic problem in low-level computer vision. Given an image with noise and/or missing pixels, the task is to obtain a 'clean' version of the image, i.e. remove the noise and fill up the missing pixels. The problem is modeled as an MRF whose variables correspond to the image pixels and whose edges define a 4-connected grid graph. The labels are the 256 possible intensity values that lie in the interval $[0, 255]$. The unary potentials are given by the squared difference between the intensity corresponding to the label and the observed intensity in the image. Since natural images are smooth, i.e. neighboring pixels tend to have similar intensity values, it is common practice to employ truncated convex pairwise potentials. In this work, we use

$$\theta_{ab}(i,j) = 30 \min\{|i - j|, 50\}. \tag{15}$$

We compared our method with the state of the art MAP estimation algorithms on two standard images, namely 'house' and 'penguin'. Fig. 4 shows the results obtained. Similar to other synthetic and real data experiments, our approach obtains labelings with lower energy values than the other move making algorithms (although it takes a longer time since it solves a series of $r$-HST metric labeling problems). In terms of the energy values, our method is outperformed by TRW-S but is computationally more efficient.

The results for scene segmentation are provided in [18].

## 7 Discussion

We presented a move making approach for the semi-metric labeling problem which approximates the given semi-metric into a mixture of $r$-HST metrics and solves each of the resulting problems using an efficient $st$-MINCUT based algorithm. Our approach provides the guarantees of the LP relaxation for the metric labeling problem. Together with the work of [5, 20], this provides further evidence of a link between randomized rounding techniques used with convex relaxations and move making algorithms. We believe that further investigations in this direction would help design move making algorithms for more complex relaxations such as [19, 28]. In practice, the results on both synthetic and real data experiments show that our method reduces the gap in performance between move making algorithms and message passing approaches. This is particularly true for applications where the unary potentials do not dominate the pairwise potentials, i.e. the prior specified by the MRF plays a vital role in obtaining good results (e.g. in scene registration). Such scenarios occur not only during testing, but during parameter learning of MRFs as well (for example, structured SVMs [31] solve a series of MAP estimation problems to learn log-linear models). An interesting direction for future research would be to generalize our move making approach to other hierarchical distance functions that approximate semi-metric distances accurately and can be learnt efficiently. Similar to the existing move making algorithms [12], the possibility of extending our approach to solve special cases of higher order potentials should also be explored.

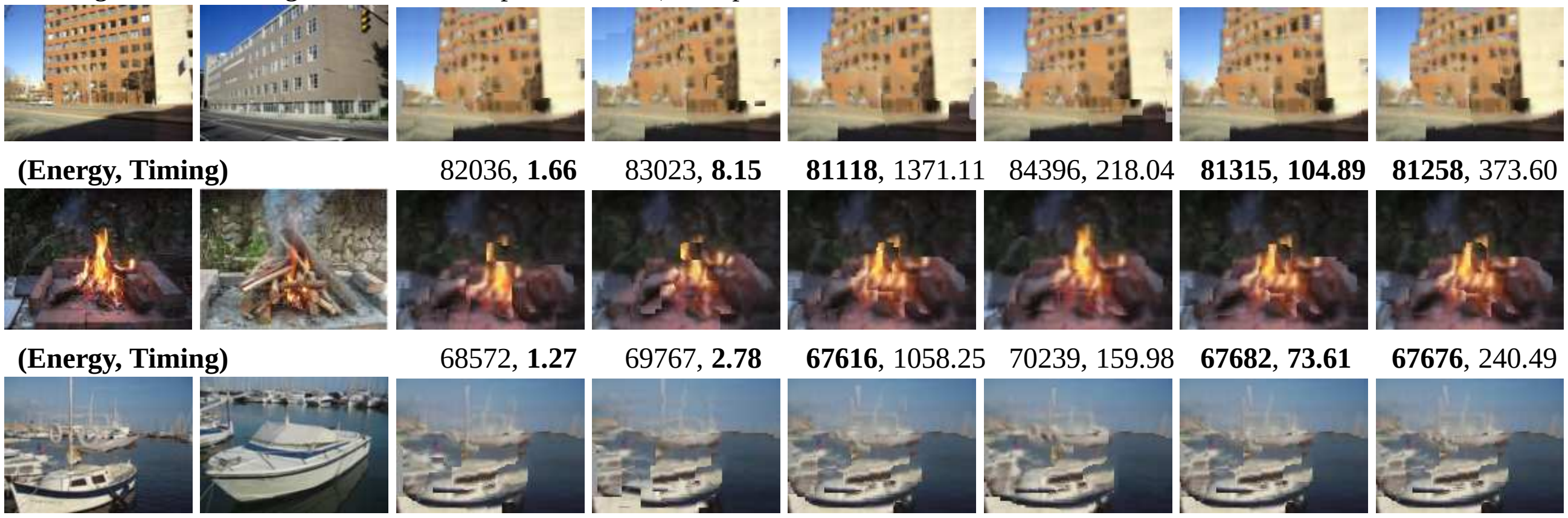

| | Image-1 | Image-2 | $\alpha$-exp | $\alpha\beta$-swap | TRW-S | BP-S | Our | Our+EM |
|---|---|---|---|---|---|---|---|---|
| **(Energy, Timing)** | | | 82036, **1.66** | 83023, **8.15** | **81118**, 1371.11 | 84396, 218.04 | **81315**, **104.89** | **81258**, 373.60 |
| **(Energy, Timing)** | | | 68572, **1.27** | 69767, **2.78** | **67616**, 1058.25 | 70239, 159.98 | **67682**, **73.61** | **67676**, 240.49 |
| **(Energy, Timing)** | | | 78222, **2.06** | 79808, **3.77** | **77457**, 1400.82 | 80002, 228.92 | **77466**, **111.88** | **77463**, 383.34 |

Figure 2: *Scene registration results. The image pairs are obtained from [22]. In each row, the first image is warped into the second image using the displacements found by various* MAP *estimation algorithms. The energy values and timings in seconds for the algorithms are shown below the corresponding warped image. The three smallest values of the energy and time required are highlighted in bold.*

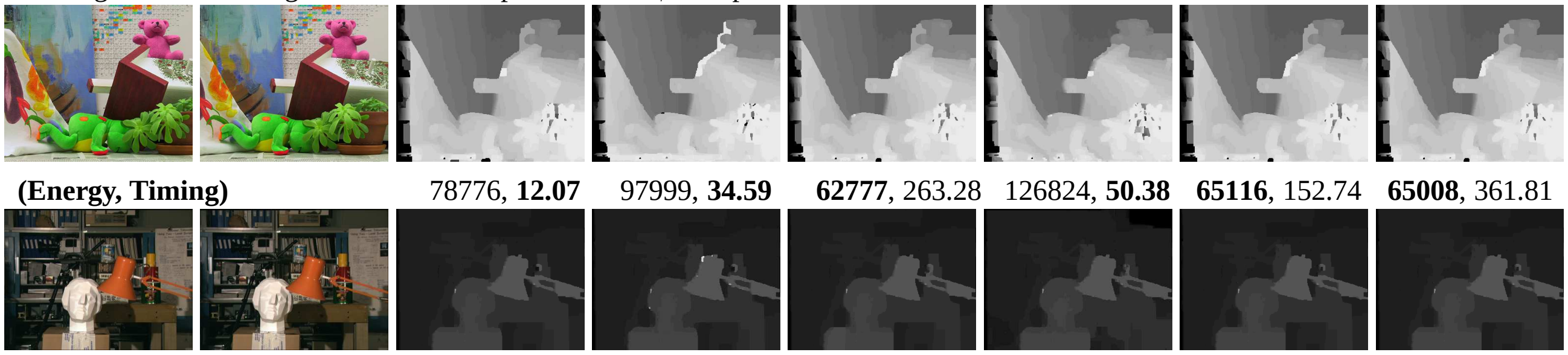

| | Image-1 | Image-2 | $\alpha$-exp | $\alpha\beta$-swap | TRW-S | BP-S | Our | Our+EM |
|---|---|---|---|---|---|---|---|---|
| **(Energy, Timing)** | | | 78776, **12.07** | 97999, **34.59** | **62777**, 263.28 | 126824, **50.38** | **65116**, 152.74 | **65008**, 361.81 |
| **(Energy, Timing)** | | | 15322, **4.49** | 18425, **13.43** | **13257**, 169.12 | 56280, **29.60** | **14135**, 72.09 | **14135**, 203.12 |

Figure 3: *Stereo reconstruction results. Each row shows the displacement map obtained by various* MAP *estimation algorithms along with their corresponding energy values and timings in seconds.*

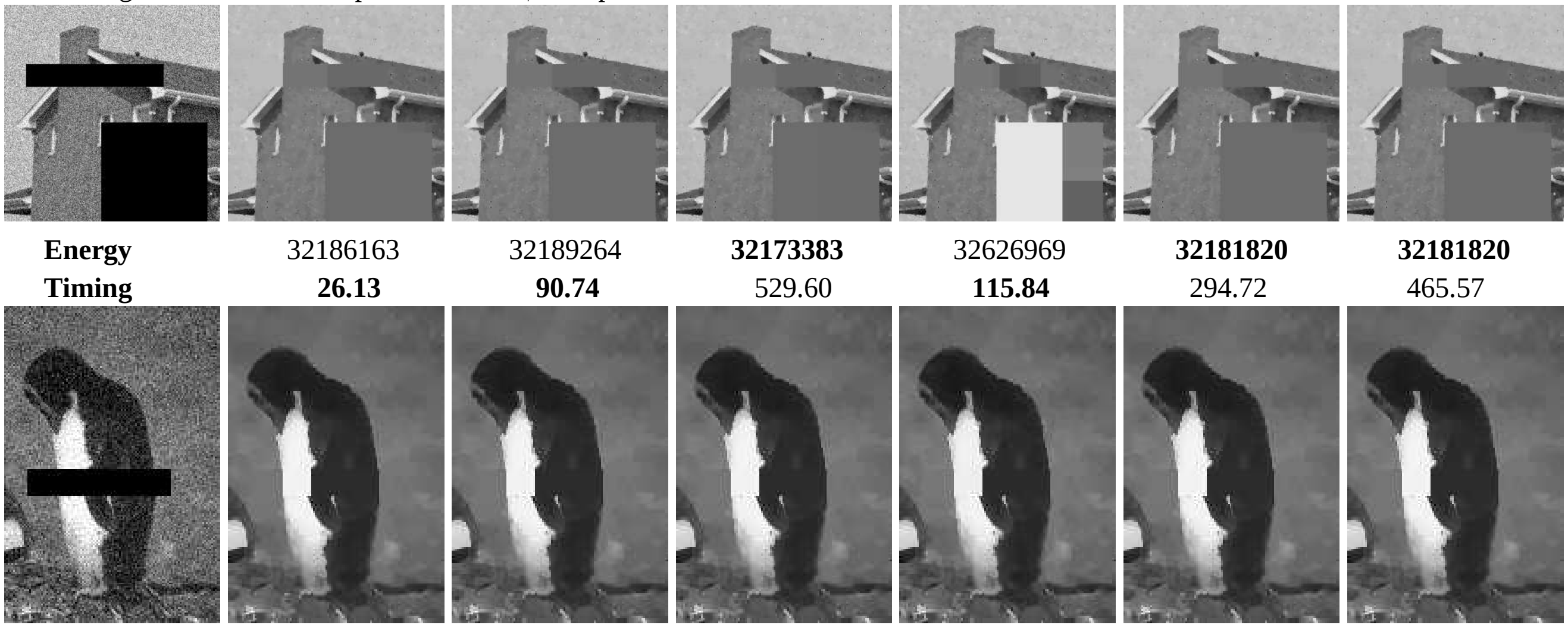

| | Image | $\alpha$-exp | $\alpha\beta$-swap | TRW-S | BP-S | Our | Our+EM |
|---|---|---|---|---|---|---|---|
| **Energy** | | 32186163 | 32189264 | **32173383** | 32626969 | **32181820** | **32181820** |
| **Timing** | | **26.13** | **90.74** | 529.60 | **115.84** | 294.72 | 465.57 |
| **Energy** | | 11075641 | 11074426 | **11068226** | 11105845 | **11072828** | **11072332** |
| **Timing** | | **5.09** | **25.22** | 174.33 | **32.94** | 70.55 | 204.55 |

Figure 4: *Image denoising results. Each row shows the 'clean' image obtained by different* MAP *estimation algorithms along with their corresponding energy values and timings in seconds. The black regions in the original image indicate missing pixels. The unary potentials for missing pixels is set to be a constant for all labels. The three lowest energy values and timings are highlighted in bold.*

**Acknowledgments.** We thank Stephen Gould for helpful discussions and careful proofreading of the manuscript. The first author is funded by the DARPA grant SA4996-10929-5.

## References

[1] Y. Bartal. On approximating arbitrary metrics by tree metrics. In *STOC*, 1998.

[2] E. Boros and P. Hammer. Pseudo-boolean optimization. *Discrete Applied Mathematics*, 123:155–225, 2002.

[3] Y. Boykov and V. Kolmogorov. An experimental comparison of min-cut/max-flow algorithms for energy minimization in vision. *PAMI*, 26(9):1124–1137, 2004.

[4] Y. Boykov, O. Veksler, and R. Zabih. Markov random fields with efficient approximations. In *CVPR*, 1998.

[5] Y. Boykov, O. Veksler, and R. Zabih. Fast approximate energy minimization via graph cuts. In *ICCV*, pages 377–384, 1999.

[6] M. Charikar, C. Chekuri, A. Goel, S. Guha, and S. Plotkin. Approximating a finite metric by a small number of tree metrics. In *FOCS*, 1998.

[7] C. Chekuri, S. Khanna, J. Naor, and L. Zosin. A linear programming formulation and approximation algorithms for the metric labelling problem. *SIAM Journal on Disc. Math.*, 18(3):606–635, 2005.

[8] J. Fakcharoenphol, S. Rao, and K. Talwar. A tight bound on approximating arbitrary metrics by tree metrics. In *STOC*, 2003.

[9] P. Felzenszwalb and D. Huttenlocher. Efficient matching of pictorial structures. In *CVPR*, pages II: 66–73, 2000.

[10] T. Hazan and A. Shashua. Convergent message-passing algorithms for inference over general graphs with convex free energy. In *UAI*, 2008.

[11] J. Kleinberg and E. Tardos. Approximation algorithms for classification problems with pairwise relationships: Metric labeling and Markov random fields. In *STOC*, 1999.

[12] P. Kohli, L. Ladicky, and P. Torr. Robust higher order potentials for enforcing label consistency. In *CVPR*, 2008.

[13] P. Kohli, A. Shekhovtsov, C. Rother, V. Kolmogorov, and P. Torr. On partial optimality in multi-label MRFs. In *ICML*, 2008.

[14] V. Kolmogorov. Convergent tree-reweighted message passing for energy minimization. *PAMI*, 2006.

[15] N. Komodakis, N. Paragios, and G. Tziritas. MRF optimization via dual decomposition: Message-passing revisited. In *ICCV*, 2007.

[16] N. Komodakis, G. Tziritas, and N. Paragios. Fast, approximately optimal solutions for single and dynamic MRFs. In *CVPR*, 2007.

[17] A. Koster, C. van Hoesel, and A. Kolen. The partial constraint satisfaction problem: Facets and lifting theorems. *Operations Research Letters*, 23(3-5):89–97, 1998.

[18] M. P. Kumar and D. Koller. MAP estimation of semi-metric MRFs via hierarchical graph cuts. Technical report, Stanford University, 2009.

[19] M. P. Kumar, V. Kolmogorov, and P. Torr. An analysis of convex relaxations for MAP estimation. In *NIPS*, 2007.

[20] M. P. Kumar and P. Torr. Improved moves for truncated convex models. In *NIPS*, 2008.

[21] V. Lempitsky, S. Roth, and C. Rother. FusionFlow: Discrete-continuous optimization for optical flow estimation. In *CVPR*, 2008.

[22] C. Liu, J. Yuen, A. Torralba, J. Sivic, and W. Freeman. SIFT flow: Dense correspondence across different scenes. In *ECCV*, 2008.

[23] D. Lowe. Object recognition from local scale-invariant features. In *ICCV*, 1999.

[24] J. Pearl. *Probabilistic Reasoning in Intelligent Systems: Networks of Plausible Inference.* Morgan Kauffman, 1998.

[25] M. Schlesinger. Syntactic analysis of two-dimensional visual signals in noisy conditions. *Kibernetika*, 1976.

[26] A. Shekhovstov, I. Kovtun, and V. Hlavac. Efficient MRF deformation model for non-rigid image matching. In *CVPR*, 2007.

[27] J. Shotton, J. Winn, C. Rother, and A. Criminisi. TextonBoost: Joint appearance, shape and context modeling for multi-class object recognition and segmentation. In *ECCV*, pages I: 1–15, 2006.

[28] D. Sontag and T. Jaakkola. New outer bounds for the marginal polytope. In *NIPS*, 2007.

[29] D. Sontag, T. Meltzer, A. Globerson, T. Jaakkola, and Y. Weiss. Tightening LP relaxations for MAP using message passing. In *UAI*, 2008.

[30] R. Szeliski, R. Zabih, D. Scharstein, O. Veksler, V. Kolmogorov, A. Agarwala, M. Tappen, and C. Rother. A comparative study of energy minimization methods for Markov random fields with smoothness-based priors. *PAMI*, 2008.

[31] I. Tsochantaridis, T. Hofmann, T. Joachims, and Y. Altun. Support vector learning for interdependent and structured output spaces. In *ICML*, 2004.

[32] O. Veksler. Graph cut based optimization for MRFs with truncated convex priors. In *CVPR*, 2007.

[33] M. Wainwright, T. Jaakkola, and A. Willsky. MAP estimation via agreement on trees: Message passing and linear programming. *Info. Th.*, 2005.

[34] Y. Weiss, C. Yanover, and T. Meltzer. MAP estimation, linear programming and belief propagation with convex free energies. In *UAI*, 2007.

[35] O. Woodford, P. Torr, I. Reid, and A. Fitzgibbon. Global stereo reconstruction under second order smoothness priors. In *CVPR*, 2008.

[36] J. Yedidia, W. Freeman, and Y. Weiss. Generalized belief propagation. In *NIPS*, pages 689–695, 2001.